\documentclass[letterpaper]{article} 
\usepackage{aaai23}  
\usepackage{times}  
\usepackage{helvet}  
\usepackage{courier}  
\usepackage[hyphens]{url}  
\usepackage{graphicx} 
\def\UrlFont{\rm}  
\usepackage{natbib}  
\usepackage{caption} 
\usepackage{algorithm}
\usepackage{algorithmic}

\usepackage{newfloat}
\usepackage{listings}
\DeclareCaptionStyle{ruled}{labelfont=normalfont,labelsep=colon,strut=off} 
\floatstyle{ruled}
\newfloat{listing}{tb}{lst}{}
\floatname{listing}{Listing}
\usepackage{booktabs}
\usepackage{amsmath}

\title{TUTORING: Instruction-Grounded Conversational Agent for Language Learners}
\author{
    Hyungjoo Chae\textsuperscript{\rm 1,3},
    Minjin Kim\textsuperscript{\rm 2},
    Chaehyeong Kim\textsuperscript{\rm 2},\\
    Wonseok Jeong\textsuperscript{\rm 3},
    Hyejoong Kim\textsuperscript{\rm 3},
    Junmyung Lee\textsuperscript{\rm 3},
    Jinyoung Yeo\textsuperscript{\rm 1,2,3}\thanks{Corresponding author.}   
}
\affiliations{
    \textsuperscript{\rm 1}Department of Computer Science, Yonsei University\\
    \textsuperscript{\rm 2}Department of Artificial Intelligence, Yonsei University\\
    \textsuperscript{\rm 3}Tutoring, Market Designers Inc.\\
    \{mapoout, minjin.kim, cheris8, jinyeo\}@yonsei.ac.kr,  ~~~ \{fredric, kate, tony\}@tutoring.co.kr
}

\newcommand{\calI}{\mbox{${\cal I}$}}
\newcommand{\calL}{\mbox{${\cal L}$}}
\newcommand{\calX}{\mbox{${\cal X}$}}

\newcommand{\ie}{{\it i.e.}}
\newcommand{\argmin}{\mathop{\mathrm{argmin}}\limits}

\begin{document}

\maketitle

\begin{abstract}
In this paper, we propose \textsc{Tutoring} bot, a generative chatbot trained on a large scale of tutor-student conversations for English-language learning. To mimic a human tutor's behavior in language education, the tutor bot leverages diverse educational instructions and grounds to each instruction as additional input context for the tutor response generation. As a single instruction generally involves multiple dialogue turns to give the student sufficient speaking practice, the tutor bot is required to monitor and capture when the current instruction should be kept or switched to the next instruction. For that, the tutor bot is learned to not only generate responses but also infer its teaching action and progress on the current conversation simultaneously by a multi-task learning scheme. Our \textsc{Tutoring} bot is deployed under a non-commercial use license at https://tutoringai.com.
\end{abstract}

\section{Introduction}\label{sec:intro}

With the recent success of neural dialogue generation~\citep{shuster2022blenderbot} based on pre-trained language models~\citep{lewis2019bart}, foreign language learning is a promising application of conversational agents in the education field. As many students experience foreign language anxiety with human tutors (called \emph{xenoglossophobia}), the conversation with AI enables the students to easily start talking in their target language. However, prior work~\citep{huang2017chatbot, pham2018chatbot, tu2020learn, shi2020language, park2021freetalky} is limited to merely ``chit-chat'', which is not thoughtfully designed for language education. In contrast, professional and dedicated human tutors may lead the conversations through diverse and personalized educational instructions, such as guiding to read sentences for beginner level, answer questions for intermediate level, or debate on controversial issues for advanced level students.

In this demo, we present a fully data-driven concept of tutoring conversational agent, which models the tutor-student conversations without any pre-defined scenario logic and manual programming for the instructions. For that, we prepare and formulate a novel dataset/task, namely \emph{instruction-grounded} response generation, where a sequence of educational instructions is described in natural language form and is shared between a tutor and a student for its use on conversations. Here, a straightforward implementation is to leverage the individual instructions as conditional code, each of which is concatenated to the dialogue context to make the desired tutor response and dialogue flow for language education. Despite its effectiveness, this approach is sub-optimal since the tutor agent cannot monitor how much conversation should be done for each instruction and whether the student has successfully followed the instruction or not.

To overcome this drawback, we design and leverage a set of auxiliary tasks that infer \emph{teaching action and progress} as dialogue state, which can be jointly learned with the primary task, \ie, instruction-grounded response generation. We hypothesize that such multi-task learning contributes to the injection of the auxiliary information into the generated tutor responses. Toward this goal, the action/progress labels of the auxiliary tasks are automatically annotated by in-dialogue signals such as human tutors' feedback, transition of the instructions, and the amount of dialogue turns per instruction. We empirically validate that such signals are effective to improve the response quality. To the best of our knowledge, beyond language models for chit-chat, \textsc{Tutoring} bot is the first instruction-based conversational agent that enables students to experience educational conversation.

\begin{figure*}[t]
\centering
    \includegraphics[width=0.95\linewidth]{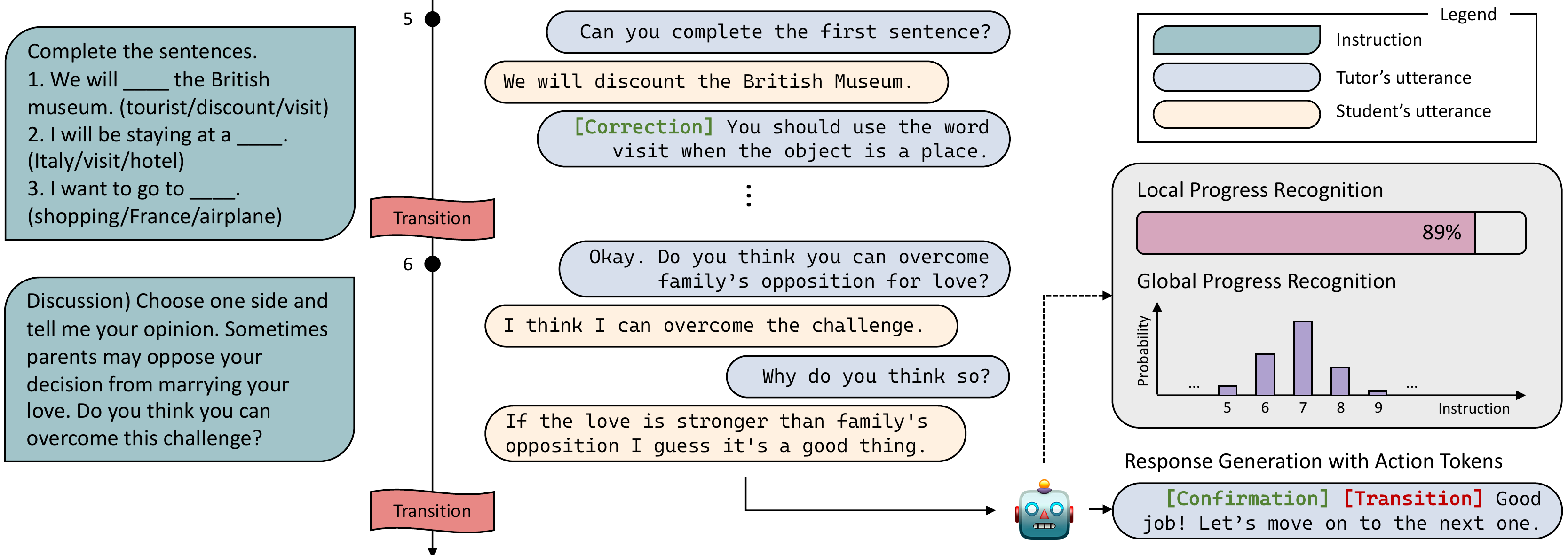}
    \caption{The system overview of \textsc{Tutoring} bot.}
    \label{fig:overview}
\end{figure*}

\section{System Description} \label{sec:system}
Figure~\ref{fig:overview} illustrates \textsc{Tutoring} bot with four auxiliary tasks of inferring action and progress information in addition to the response generation task based on multi-task learning.

\subsection{Tutor Response Generation with Action Codes}\label{sec:system.1}

Let $\calX$ and $\{\calI_i\}_{i=1}^N$ be a dialogue context with $T$ turns and a fixed sequence of $N$ instructions, where each turn is aligned with one specific instruction $\calI_i$. A dialogue model parameterized by $\theta$ aims to generate an appropriate response $y$ with $M$ tokens based on $\calX$ and $\calI_i$.

To allow the model to ground on instructions, we introduce action codes for the instruction and dialogue, respectively. The instruction action code $y^{inst}$ indicates whether to move on to the next instruction $\calI_{i+1}$, which is a special token \texttt{[Transition]} generated only when the conversation ends for the current instruction $\calI_i$. On the other hand, the dialogue action code $y^{dial}$ represents educational feedback for the given dialogue context $\calX$, which can be either \texttt{[Correction]}, \texttt{[Confirmation]}, or \texttt{[Others]}.

Given a dialogue context $\calX$ and its aligned instruction $\calI_i$, the model learns to generate a response $y$ with the action codes $y^{inst}$ and $y^{dial}$. The generation loss $\calL_{gen}$ is computed by the negative log-likelihood loss.
\begin{multline}
    \calL_{gen} = - \{\log p(y^{dial}|\calX,\calI_i) + \log p(y^{inst}|\calX,\calI_i,y^{dial}) \\ + \sum_{j=1} ^{M} \log p(y_j|\calX,\calI_i,y^{dial}, y^{inst}, y_{<j}) \}
\end{multline}

\subsection{Global and Local Progress Recognition}\label{sec:system.2}

Considering the sequence of instructions from $\{\calI_i\}_{i=1}^N$, we design the global progress recognition task, where the global progress $y^{glo}_{rec}$ represents which instruction is involved in the current conversation (\ie, the index $i$ from $\calI_i$). We further incorporate the local progress recognition task, in which the local progress $y^{loc}_{rec}$ denotes the fraction of the number of proceeded dialogue turns over the total number of dialogue turns aligned with $\calI_i$, thus ranging from 0 to 1.

The model learns to predict both global and local progress by the recognition loss $\calL_{rec}$, which is defined as the sum of cross-entropy and mean squared error loss between ground truth labels $y_{rec}$ and predicted labels $\hat{y}_{rec}$, respectively.
\begin{equation}
    \calL_{rec} = \texttt{CE}(y^{glo}_{rec}, \hat{y}^{glo}_{rec}) + \texttt{MSE}(y^{loc}_{rec}, \hat{y}^{loc}_{rec})
\end{equation}

\subsection{Multi-task Learning}\label{sec:system.3}
The dialogue model is jointly trained on the aforementioned tasks to generate instruction-grounded responses and recognize the learning progress by updating $\theta$ as follows:
\begin{equation}
    \textsc{Tutoring}~\text{bot}: \theta^* = \argmin_\theta ~ \calL_{gen}(\theta) + \calL_{rec}(\theta)
\end{equation}

\section{Evaluation and Demonstration}

Based on 11 unique instructions, we collect 1,911 dialogues of 95,343 utterances from tutor-student conversations in real world. We split the dataset with ratio 8:1:1 for the training, validation, and test sets, respectively, where the instructions are shared between the sets. We employ a pre-trained BART-large~\citep{lewis2019bart} as our base model. The evaluation results in Table~\ref{tab:auto_eval} show that incorporating both action codes and progress recognition achieves the best performance with their synergistic effect in the response generation task. Also, instruction transitions achieve 87.98\% in terms of accuracy.

\begin{table}[tbp]\small
    \centering
    \begin{tabular}{l cccc}
        \toprule 
        Model & BLEU-1 & BLEU-2 & BLEU-3 & BLEU-4 \\
        \midrule 
        RG & 21.03 & 14.42 & 9.57 & 6.95 \\
        RG + AC & 21.44	& 14.70 & 9.78 & 7.14 \\
        RG + PR & 21.53 & 14.73 & 9.79 & 7.14 \\
        RG + AC + PR & \textbf{23.02} & \textbf{15.78} & \textbf{10.59} & \textbf{7.74} \\
        \bottomrule
    \end{tabular}
    \caption{Performance of the proposed model under different configurations. RG, AC, and PR denotes response generation, action codes, and progress recognition, respectively.}
    \label{tab:auto_eval}
\end{table}

We deploy \textsc{Tutoring} bot to online education demo by attaching Speech-to-Text\footnote{Vosk Speech Toolkit: https://github.com/alphacep/vosk-api} and Text-to-Speech\footnote{Coqui Tacotron2-DCA: https://github.com/coqui-ai/TTS} modules for making the system easily accessible. The expected running time of a tutoring session is about 11 minutes with 50 turns on average. Our demo video shows that \textsc{Tutoring} bot can properly control the real-time teaching process and timely transition within instructions with proper educational feedback. We additionally provide a debugging tool for developers to identify the generated action codes and the results of the progress recognition tasks based on visualization using Gradio~\citep{abid2019gradio}.

As future work, we will release a public and full version of our task and dataset with 50 unique instructions and 922,446 utterances soon.
Using this dataset as a testbed, the tutor bot can be advanced to a product-ready application with 1) more generalized dialogue models for diverse/unseen instructions, 2) extended auxiliary tasks for better response quality, and 3) additional feedback/correction modules on expressiveness, grammar, and pronunciation.

\section{Acknowledgments}
We would like to thank anonymous reviewers for their valuable comments. This work was partially supported by the Institute of Information \& communications Technology Planning \& Evaluation (IITP) grant funded by the Korea government (MSIT) (No. 2020-0-01361, Artificial Intelligence Graduate School Program (Yonsei University)) and the National Research Foundation of Korea (NRF) grant funded by the Korea government (MSIT) (No. 2022-11-0941).
\bibliography{aaai23}

\end{document}